\documentclass{article}

\usepackage{arxiv}

\usepackage[utf8]{inputenc} 
\usepackage[T1]{fontenc}    
\usepackage{hyperref}       
\usepackage{url}            
\usepackage{booktabs}       
\usepackage{amsfonts}       
\usepackage{nicefrac}       
\usepackage{microtype}      
\usepackage{lipsum}		
\usepackage{graphicx}
\usepackage{natbib}
\usepackage{doi}
\usepackage{dirtytalk}
\usepackage[table,x11names]{xcolor}
\usepackage{amsmath}
\usepackage{booktabs}
\usepackage{adjustbox}
\usepackage{amsthm}
\usepackage{colortbl}
\usepackage{multirow}

\title{Multimodal HIE Lesion Segmentation in Neonates: A Comparative Study of Loss Functions }

\author{
    {\hspace{1mm}Annayah Usman,}
    {\hspace{1mm}Abdul Haseeb,}
    \href{https://orcid.org/0000-0003-0638-9689}{\includegraphics[scale=0.06]{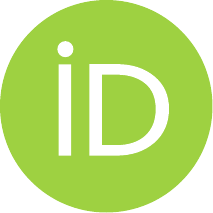}\hspace{1mm}Tahir Syed} \\
    School of Mathematics and Computer Science, Institute of Business Administration Karachi, Pakistan \\
    \texttt{\{tahirqsyed\}@gmail.com} }

\renewcommand{\shorttitle}{\textit{MatAE}  here goes short title}

\hypersetup{
pdftitle={A template for the arxiv style},
}

\begin{document}
\maketitle

\begin{abstract}

Segmentation of Hypoxic-Ischemic Encephalopathy (HIE) lesions in neonatal MRI is a crucial but challenging task due to diffuse multifocal lesions with varying volumes and the limited availability of annotated HIE lesion datasets. Using the BONBID-HIE dataset, we implemented a 3D U-Net with optimized preprocessing, augmentation, and training strategies to overcome data constraints. The goal of this study is to identify the optimal loss function specifically for the HIE lesion segmentation task. To this end, we evaluated various loss functions, including Dice, Dice-Focal, Tversky, Hausdorff Distance (HausdorffDT) Loss, and two proposed compound losses—Dice-Focal-HausdorffDT and Tversky-HausdorffDT—to enhance segmentation performance. The results show that different loss functions predict distinct segmentation masks, with compound losses outperforming standalone losses. Tversky-HausdorffDT Loss achieves the highest Dice and Normalized Surface Dice scores, while Dice-Focal-HausdorffDT Loss minimizes Mean Surface Distance. This work underscores the significance of task-specific loss function optimization, demonstrating that combining region-based and boundary-aware losses leads to more accurate HIE lesion segmentation, even with limited training data.
\end{abstract}

\section{Introduction}
\label{sec: intr}
Hypoxic-Ischemic Encephalopathy (HIE) refers to a brain injury caused by reduced oxygen or blood flow during the prenatal, intrapartum or postnatal period \cite{NINDSn.d.}. With an incidence rate of 1 to 5 per 1,000 term births, HIE affects approximately 750,000 term-born newborns worldwide each year, making it a major cause of pediatric mortality and morbidity, imposing a significant economic burden of $\$ 2 $ billion annually in the United States alone \cite{Bao2023,graham2008systematic,lee2013intrapartum}. Beyond its high prevalence and substantial economic burden, HIE has severe health implications, with up to $60 \%$ affected infants succumbing or developing neurocognitive impairments such as mental retardation, epilepsy, and cerebral palsy \cite{Allen2011}. Although therapeutic hypothermia, the current clinical treatment of HIE, can reduce the risk of death or disability in infants with moderate or severe HIE \cite{shankaran2005whole}, more than 1/3 of patients still die or suffer with moderate to severe disability \cite{shankaran2012brain}.

Magnetic Resonance Imaging (MRI) is predominantly used in HIE-related clinical trials \cite {weiss2019mining}. Early MRI is particularly effective in detecting severe brain injury and provides a consistent prediction of adverse outcomes \cite{okane2021early,thoresen2021mri}. In addition, specific brain injury patterns represented by MRI lesions serve as key markers of death or disability \cite{shankaran2012brain}. Therefore, the precise identification of brain lesions in neonatal MRI is crucial for disease prognosis and evaluation of treatment effects, improving neonatal care and survival rates \cite{hung2024mri,rutherford2010magnetic}.

Although advances in Deep Learning (DL) have enabled accurate segmentation of large and focal brain lesions, such as those of brain tumors, state-of-the-art DL models struggle with segmentation of HIE lesions as they are typically multi-focal, diffused, and often occupy a very small proportion of brain volume (< $1 \%$), making them difficult to segment accurately \cite{Bao2023}. Additionally, limited dataset availability prevents effective use of advanced transformer-based architectures, optimal hyper-parameter tuning, and invariance to scanning protocols, leading to issues like over-fitting and poor generalizability.

Segmentation performance in medical imaging is highly dependent on the choice of loss function, as different loss functions produce distinct segmentation maps, each suited to specific tasks \cite{ma2021loss,delgado2024advancing}. Several comparative studies have analyzed different loss functions in medical imaging, each designed to tackle specific challenges. Some focus on data imbalance, addressing disparities between background and foreground as well as easy and hard examples, while others emphasize structural accuracy, such as precise boundary segmentation in difficult regions. Additionally, certain loss functions aim to balance false positives and false negatives, reducing over- and under-segmentation \cite{kato2023adaptive,zhao2020rethinking,yeung2021unified,jadon2020survey}. Among these, compound loss functions have demonstrated greater robustness across tasks \cite{ma2021loss}. However, no study has systematically evaluated the impact of loss functions on HIE lesion segmentation, highlighting a critical gap.

Reviewing recent literature reveals work around ensemble approaches, heavy data augmentations, and advanced architectures such as Swin-UNetR, nnUNet, SageResNet, and fusion models (e.g., Swin-UNetR with a random forest classifier) \cite{toubal2025fusion,kazemi2025enhancing,koirala2025ensemble,wodzinski2025improving,tahmasebi2025deep,aydin2025segresnet}. Some of these studies have experimented with different compound losses while others have just used Dice loss, with differences in pre-processing and modelling architectures. The absence of a controlled setting and a consistent baseline makes it difficult to isolate the impact of loss functions on the segmentation performance. In this study, we aim to identify the most suitable loss function for HIE lesion segmentation by systematically comparing different loss functions under a controlled setting. Our findings will contribute to the BONBID-HIE challenge \cite{GrandChallenge2024} and provide valuable insights for researchers working on HIE lesion segmentation, enabling them to combine advanced architectures with sophisticated compound loss function suitable for HIE lesions particularly to further improve segmentation precision.

\section{Loss Functions}
Loss functions for medical image segmentation can be broadly categorized into four types: region-based, distribution-based, boundary-based, and compound loss functions \cite{ma2021loss}. Region-based loss functions aim to maximize the overlap between the ground truth and predicted segmentation masks. Distribution-based loss functions minimize the dissimilarity between the predicted and actual probability distributions. Boundary-based loss functions focus on minimizing the distance between the predicted and ground truth boundaries, ensuring precise delineation of structures. Lastly, compound loss functions combine the strengths of the previous three types through a weighted formulation, leveraging their complementary strengths while offsetting their individual downsides.

In this study, we evaluate a diverse set of loss functions specifically for multimodal HIE lesion segmentation in neonates. Our selection is motivated by the need to balance region-based overlap, boundary precision, and class imbalance handling. Given these considerations:
\begin{itemize}
    \item \textbf{Dice Loss} \cite{Milletari2016} serves as the baseline, being one of the most widely adopted loss functions for medical image segmentation \cite{kato2023adaptive}.
\end{itemize}
\begin{itemize}
    \item \textbf{Tversky Loss} \cite{Sadegh2017}, an extension of Dice Loss, is employed to better handle imbalanced lesion sizes, which is particularly relevant to our case.
\end{itemize}
\begin{itemize}
    \item \textbf{Hausdorff Distance Loss} \cite{Karimi2019} is incorporated to explicitly encourage accurate boundary delineation.
\end{itemize}
\begin{itemize}
    \item Compound Losses are explored to integrate the strengths of different loss types while mitigating individual limitations. Specifically, we evaluate:
    \begin{itemize}
        \item \textbf{Dice-Focal Loss} \cite{Zhu2018AnatomyNet}, which combines region-based (Dice) and distribution-based (Focal) components to enhance segmentation in cases of class imbalance.
    \end{itemize}
    \begin{itemize}
        \item \textbf{Dice-Focal-Hausdorff Distance Loss} and \textbf{Tversky-Hausdorff Distance Loss} are new compound loss functions defined, which further incorporate boundary-aware constraints to improve segmentation precision.
    \end{itemize}
\end{itemize}

Distribution-based losses excel at pixel-wise classification, which might classify the majority of background pixels correctly; however, they do not account for spatial coherence, which leads to fragmented masks and fails to capture precise structures. Hence, we have not used them as standalone, as in lesion segmentation, region consistency and boundary precision are more critical than per-pixel classification. However, we incorporate Focal Loss (distribution-based) within compound losses (e.g., Dice-Focal, Dice-Focal-Hausdorff) to leverage its strength in handling class imbalance, without using it independently.\\

\noindent The following are the detailed definitions of the losses used:

\begin{itemize}
  \item \textbf{Dice Loss:} It's used to measure the similarity between two sets.
\end{itemize}
\begin{equation*}
\text { DiceLoss }=1-\frac{2|A \cap B|}{|A|+|B|} \tag{1}
\end{equation*}

 Where, A and B are the predicted and ground truth binary masks, respectively.

\begin{itemize}
  \item \textbf{Dice Focal Loss (DFL):} Dice Focal Loss combines both the Dice Loss and Focal Loss, focusing on hard-to-classify examples.
\end{itemize}
\begin{equation*}
D F L=(1-\alpha)(1-\text { DiceLoss })+\alpha(1-\text { FocalLoss });\text { FocalLoss }=-\lambda_{t}\left(1-p_{t}\right)^{\gamma} \log \left(p_{t}\right) \tag{2}
\end{equation*}

 Where $\alpha$ is a balancing factor between Dice loss and Focal loss, $p_{t}$ is predicted probability mask of the true class, $\lambda_{t}$ is a weighting factor for the class, and $\gamma$ is the focusing parameter that adjusts the rate at which easy examples are down-weighted. By default, $\gamma$ is kept as 2, equal balance is given to both Dice and Focal loss, and no weight is applied.

\begin{itemize}
  \item \textbf{Tversky Loss:} Tversky Loss is a generalization of Dice loss and is useful in dealing with class imbalances, particularly when false positives and false negatives need to be weighted differently.
\end{itemize}

\begin{equation*}
\text { TverskyLoss }=1-\frac{|A \cap B|}{|A \cap B|+\alpha|\sim B|+\beta|\sim A|}=1-\frac{T P}{T P+\alpha F P+\beta F N} \tag{3}
\end{equation*}

\noindent Where A and B are same as Dice loss and $\alpha$ and $\beta$ are hyperparameters that control the weighting of false positives and false negatives. We've kept $\alpha$ as 0.3 and $\beta$ as 0.7 $(\alpha<\beta)$ to emphasize of false negatives more as in medical diagnosis missing a lesion could have severe consequences than wrongly identifying one.

\begin{itemize}
  \item \textbf{Hausdorff Distance Loss (HDTL):} Hausdorff Distance is a measure of the maximum distance between two sets. In segmentation, it's used to measure how far apart the predicted and true boundaries are. Hausdorff Distance Loss is the reciprocal of this value to minimize the distance:
\end{itemize}

\begin{equation*}
H D T L=\frac{1}{1+\text { HausdorffDistance }} \tag{4}
\end{equation*}

\begin{itemize}
    \item \textbf{Dice-Focal-Hausdorff Distance Loss and Tversky-Hausdorff Distance Loss:} We also defined two new compound loss functions. One is a linear combination of Dice-Focal Loss and Hausdorff Distance Loss, while the other is a linear combination of Tversky Loss and Hausdorff Distance Loss. Through the first combination, the aim is to strengthen Dice-Focal Loss's ability to segment lesions of varying sizes by incorporating Hausdorff Distance Loss for improved boundary precision. In the second combination, the goal is to merge Tversky Loss's strength in capturing large lesions with Hausdorff Distance Loss's ability to capture precise boundaries. They're defined as below:
\end{itemize}

\begin{equation*}
DF-HDTLoss=\alpha(DFL)+\beta(\log (HDTL)) \tag{5}
\end{equation*}
\begin{equation*}
Tversky-HDTLoss=\alpha(TverskyLoss)+\beta(\log (HDTL)) \tag{6}
\end{equation*}

\noindent Taking the log of the Hausdorff distance loss in both combinations reduces the impact of outlier distances, stabilizing training and emphasizing overall boundary alignment rather than rare extreme errors. We've kept $\alpha$ as 0.9 and $\beta$ as 0.1 after empirically testing combinations from $\{(\alpha, \beta): \alpha \in[0.5,0.9], \beta \in[0.1,0.5]\}$. All of the above, as well as the evaluation metrics described ahead, are implemented through MONAI library \cite{MONAIhome}.

\section{Experiments}

\subsection{Dataset and Benchmarking}
\textbf{Dataset.} The dataset used here is Part I of BONBID-HIE 2024 Grand Challenge \cite{GrandChallenge2024}. It contains 3d skull stripped Apparent Diffusion Coefficient (ADC) maps, Z-score normalized ADC maps (ZADC), and binary label masks for 133 HIE patients ( 85 training cases, 4 validation cases, and 44 testing cases - hidden). ADC maps measure water diffusion in brain tissues, with restricted movement suggesting lesions. ZADC maps are a normalized version of ADC maps showing deviation from normal range. Normal values were constructed from ADC maps of 13 healthy individuals acquired $0-14$ days after birth \cite{Bao2023}. This dataset is particularly challenging because while it is the first MRI dataset specifically created for HIE segmentation tasks, it is extremely small to utilize advanced architectures properly, it has lesion volumes that vary greatly from $<1 \%$ to $50-100 \%$, and its maps are susceptible to scanner variances.

\noindent \textbf{Benchmark.} Segmentation accuracy, measured by Dice overlap with state-of-the-art deep learning models such as UNet \cite{Ronneberger2015} on HIE lesions, remains around 0.5 \cite{Bao2023,murphy2017automatic}.

\subsection{Experimental Setup}

\subsubsection{Experimental Design}
The experimental design follows a comparative study framework to systematically evaluate the impact of different loss functions on HIE lesion segmentation. To ensure a controlled setting, the dataset is preprocessed and augmented identically, and the segmentation model is fixed across all loss function experiments to isolate their effect on segmentation precision. The study aims to optimize the loss function specifically for HIE lesions, with Dice loss serving as the baseline. Performance is assessed using standard segmentation metrics—Dice Score \cite{dice1945measures}, Mean Surface Distance (MSD) \cite{benes2015performance}, and Normalized Surface Dice (NSD) Score \cite{nikolov2021deep}. The goal is to determine whether the proposed loss functions surpass the baseline and exceed existing benchmarks for HIE lesion segmentation.
\begin{figure*}[h]
    \centering
    \includegraphics[width=\textwidth, height=\textwidth, keepaspectratio]{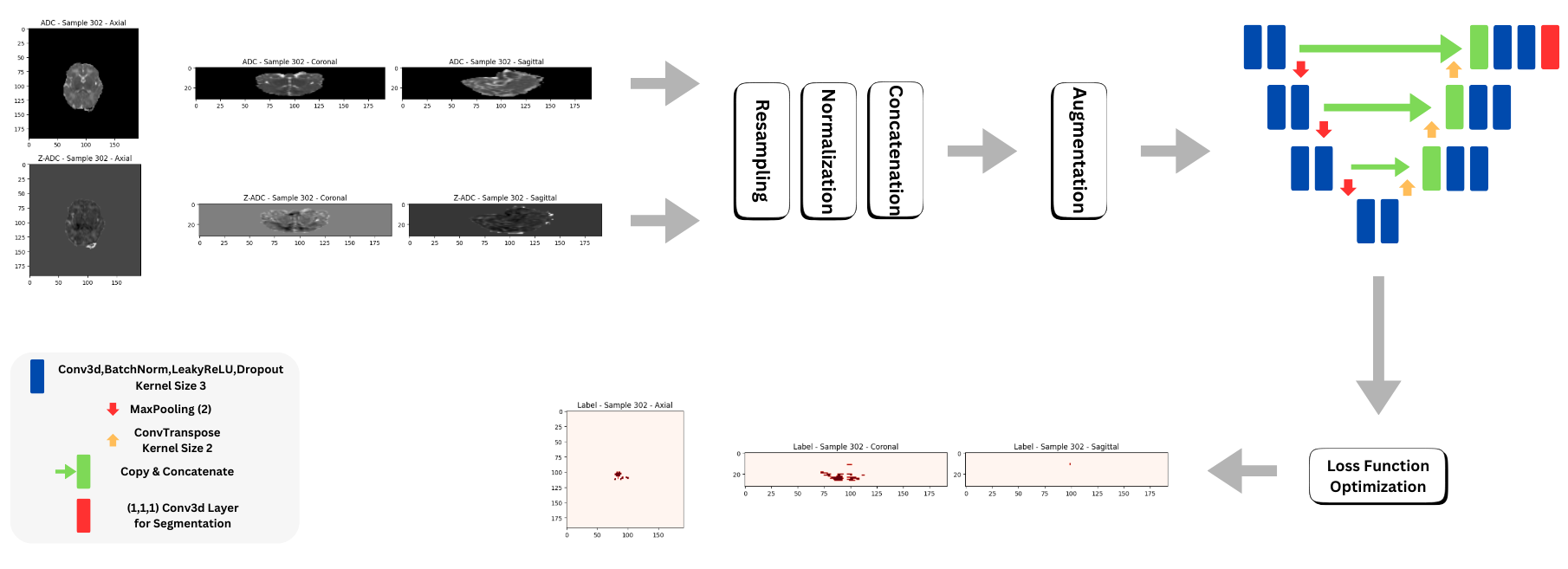}
    \caption{\textbf{Overview of the Experimental Setup.} The setup processes two input modalities: 3D ADC and ZADC maps, resampling them with trilinear interpolation and normalizing their intensities separately. The concatenated 2-channel 3D input (2, 192, 192, 32) then undergoes various data augmentations. A U-Net with three encoder-decoder blocks is trained using different loss functions, and segmentation performance is evaluated using the Dice Score, MSD, and NSD Score.}
\end{figure*}

\subsubsection{Model Architecture and Training Protocol}
We've specified a 3D UNet architecture, due to its strong inductive bias, instead of employing more recent architectures like ViTs, which generally require larger datasets to realize their advantages. Our design includes three encoder and decoder blocks instead of the conventional four, reducing depth and parameter count to prevent overfitting given the limited dataset size. Each encoder block consists of double 3D convolutional (conv) layers followed by batch normalization, LeakyReLU activation, and dropout, concluding with a maxpooling layer for down-sampling. The bottleneck further enhances feature representations through additional double conv layers. The decoder progressively up-samples using transpose convolutions, integrates encoder features via skip connections, and then applies double conv layers. The output layer employs a $(1,1,1)$ convolution to generate the segmentation map.

We've used complete 2-channel input, combining ADC and ZADC maps, rather than single channel patches to leverage global context and multimodal information. Batch normalization ensures faster and more stable learning, while dropout minimizes overfitting, with progressively higher rates in deeper layers to encourage independent feature learning across activation maps at higher levels. LeakyReLU maintains a small gradient for negative inputs, ensuring gradient flow through non-lesion areas and enabling the model to learn from both lesion and background regions, which is crucial for sparse lesion segmentation.\\

\noindent \textbf{Preprocessing.} 3D medical images are memory intensive and so to reduce computational overhead during training, the data is preprocessed offline as doing it on-the-fly can lead to duplicate memory allocation. The following three preprocessing steps are undertaken:
\begin{itemize}
        \item ADC and ZADC maps and label masks are resampled to a fixed size to make Pytorch batching possible and help model learn efficiently without being biased by size variations. Maps and labels are resampled to $(192,192,32)$ which is maximum of average dimension across all directories, rounded to a multiple of $2^{\wedge} 4=16$ (total down-sampling factor of traditional UNet) to ensure clean down-sampling and up-sampling (without fractional sizes that require inconsistent padding). For maps resizing is done using Trilinear interpolation to provide smooth transition in intensity values and for mask Nearest-Neighbor interpolation is used because it preserves discrete class boundaries. We didn't resample to maximum dimension across all directories purely due to memory constraints as otherwise we believe it to be the superior choice.
        \item The resampled (resized) maps' intensities are normalized individually using mean and standard deviation to make the model more robust by mitigating differences in scanners and patient anatomy.
        \item Lastly, both ADC and ZADC maps are concatenated (stacked on top of each other) to make a 2-channel input $(2,192,192,32)$ for UNet to allow the model to leverage information from both modalities.
\end{itemize} 

\noindent \textbf{Data Augmentation.} During training, the maps are augmented using the TorchIO library \cite{TorchIO} with a probability of 0.5 for each transformation: (i) Random Noise (mean $=0.0$, std=0.01) simulates realistic scanner artifacts, (ii) Random Anisotropy (downsampling=(1.2, 2.0)) reflects mild-to-moderate real-world variations in image resolution, (iii) Random Blur ((std=(0, 0.5)) addresses scenarios where lesion edges appear blurred due to low resolution or motion artifacts, (iv) Random Gamma ( $\log \_$gamma $=(-0.1,0.1)$ ) introduces subtle realistic changes to brightness and contrast, and (v) Random Elastic Transformation mimics natural variations in soft tissues caused by patient movement or anatomy. Augmentations are crucial in our case to artificially increase the training dataset size, simulate real world imaging variations for improved generalizability, and ensure scanner invariance in the model. \\

\noindent \textbf{Machine Specification and Hyperparameter Selection.} The network is trained on Nvidia Tesla P100 GPU on Kaggle with a batch size of 4, the maximum feasible size given memory constraints. Training is capped at 100 epochs, with early stopping enabled to halt training after 10 epochs with no improvement. To improve generalization and prevent overfitting due to the limited dataset size, the AdamW optimizer is used. Unlike the original Adam where weight decay is implicitly tied to the learning rate, AdamW explicitly applies weight decay by subtracting weight penalty during parameter update, which encourages smaller weights and smoother decision boundaries. A weight decay rate of $1 \mathrm{e}-3$ complements the small batch size by countering the risk of overfitting due to noisy gradient updates. The learning rate is set to $1 \mathrm{e}-3$, with exponential decay (factor of 0.9 ) for faster convergence and reduced oscillations in later epochs. L1 regularization (1e-4) promotes sparsity in the model, reducing overfitting. Gradient clipping ( $\max$ norm $=1$ ) is applied to prevent gradient explosion, which is crucial for deep architectures like UNet. These strategies-dropout, weight decay, L1 regularization, and gradient clipping-help stabilize training, reduce overfitting, and enhance generalizability. \\

\noindent Note: We've used Pytorch \cite{PyTorch} library mainly for the implementation of this study. 

\subsection{Evaluation Metrics}
For the comparison of segmentation quality across different loss functions, we evaluated them using Dice Score, Mean Surface Distance (MSD), and Normalized Surface Dice (NSD) Score. The use of Dice, NSD, and MSD as evaluation metrics is well justified based on both theoretical and empirical evidence. Prior work \cite{eelbode2020optimization} has shown that Dice and Jaccard scores remain the best indicators of segmentation performance when metric-sensitive losses are used, such as Dice and Tversky, because they optimize the same overlap-based criteria used for evaluation, assessing model’s performance in line with its training objectives. Our study builds upon this by incorporating both overlap-based (Dice) and boundary-sensitive (NSD, MSD) metrics to comprehensively evaluate the segmentation performance of both overlap-based and boundary-aware loss functions. Given the challenges of capturing small HIE lesions and ensuring accurate boundary delineation, these metrics provide the most appropriate evaluation framework for our study. Additionally, these metrics align with standard evaluation protocols in medical imaging and were specifically required by the BONBID-HIE Grand Challenge organizers, further supporting their validity in this domain. 

\noindent These metrics are defined below:

\begin{itemize}
  \item \textbf{Mean Surface Distance (MSD):}
\end{itemize}

\begin{equation*}
\operatorname{MSD}(p, q)=\frac{1}{2}\left(\frac{d(\delta(q), \delta(p))}{|\delta(q)|}+\frac{d(\delta(p), \delta(q))}{|\delta(p)|}\right) \tag{7}
\end{equation*}

\noindent Where $\delta(x)$ represents the surface of $x$ and $d(x, y)$ represents the shortest surface distance from surface $x$ to surface $y$. It computes the average distance between the surfaces of two binary masks, measuring how well the predicted and ground truth surfaces align. It considers both directions: from the prediction surface to the ground truth and vice versa.

\begin{itemize}
  \item \textbf{Normalized Surface Dice (NSD):}
\end{itemize}

\begin{equation*}
N S D(p, q)_{\tau}=\frac{\left|\delta(q) \cap \gamma_{\tau}(p)\right|+\left|\gamma_{\tau}(q) \cap \delta(p)\right|}{|\delta(q)+\delta(p)|} \tag{8}
\end{equation*}

\noindent NSD is a metric used to evaluate the similarity between the boundary surfaces of predicted and ground truth binary masks, allowing for a specified tolerance distance $\tau$. It considers how many boundary points from the two masks fall within this distance, rather than requiring exact overlap. NSD provides a more flexible, surface-focused measure of accuracy, especially useful in scenarios where small boundary mismatches are acceptable.

\begin{itemize}
  \item \textbf{Dice Coefficient:}
\end{itemize}

\begin{equation*}
\operatorname{Dice}(p, q)=\frac{2 \times|p \cap q|}{|p|+|q|} \tag{9}
\end{equation*}

\noindent It is a measure of the volumetric overlap between predicted and ground truth binary masks relative to their total size.

\subsection{Results and Discussion}
\begin{table}[!ht]
\centering
\begin{tabular}{|c|c|c|c|c|}
\hline
\textbf{Loss Functions} & \textbf{Dice $\uparrow$} & \textbf{MSD $\downarrow$} & \textbf{NSD $\uparrow$} & \textbf{Epochs} \\
\hline
Dice Loss (Baseline) & 0.3800 & 15.0650 & 0.3850 & 32 \\
\hline
Dice Focal Loss & 0.4900 & 1.7925 & 0.5275 & 49 \\
\hline
Tversky Loss & 0.3525 & 15.3650 & 0.3375 & 38 \\
\hline
HausdorffDT Loss & 0.3300 & Inf & 0.2800 & 29 \\
\hline
\begin{tabular}{c}
DiceFocal-HausdorffDT \\
Loss \\
\end{tabular} & 0.4925 & $\mathbf{1 . 4 2 2 5}$ & 0.5300 & 72 \\
\hline
\begin{tabular}{c}
Tversky-HausdorffDT Loss \\
\end{tabular} & $\mathbf{0 . 5 0 0 0}$ & 1.6250 & $\mathbf{0 . 5 3 2 5}$ & 59 \\
\hline
\end{tabular}
\caption{Dice, MSD, and NSD scores of different loss functions. Bold values, other than headers, correspond to best scores.}
\label{tab:loss_comparison}
\end{table}

\begin{figure}[h]
    \centering
    \includegraphics[width=\textwidth]{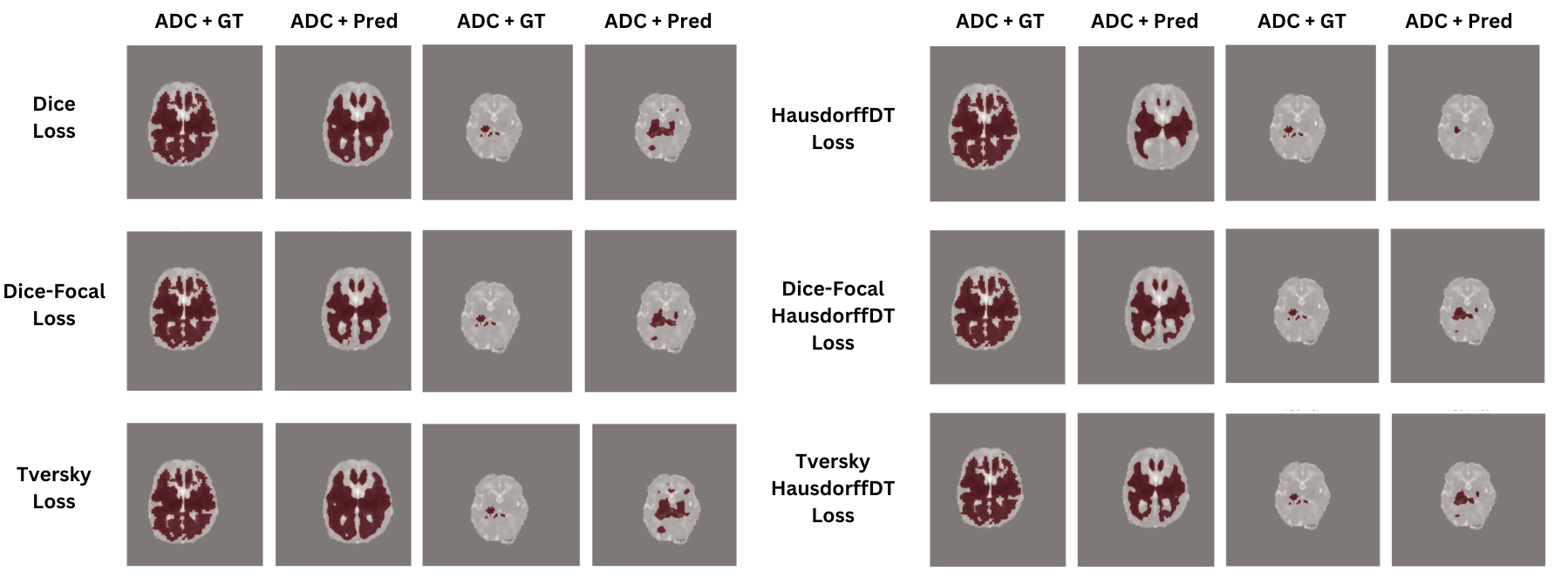}
    \caption{Visual evaluation of predicted segmentation masks versus ground truth on Axial ADC maps across all loss functions, focusing on both extremely small and large lesion types for a holistic performance assessment.}
    \label{fig:segmentation_results}
\end{figure}

The results indicate that studying and optimizing task-appropriate loss functions is crucial. As shown in Figure~\ref{fig:segmentation_results}, the segmentation masks produced by different loss functions exhibit distinct patterns, highlighting their varying impacts on lesion segmentation. Moreover, combining loss functions yields superior performance compared to standalone loss functions. For instance, Dice-Focal Loss shows a significant improvement over the baseline Dice Loss. This is expected, as Dice Loss focuses on overlap between predicted and ground truth regions, while Focal Loss specifically prioritizes hard-to-segment regions, such as smaller or less prominent lesions, over larger, well-defined ones. This targeted approach leads to better overall segmentation performance.

Similarly, Hausdorff Distance Loss aims to improve boundary precision; however, it often fails to capture complete lesion areas and sometimes misses small lesions entirely. In contrast, Tversky Loss performs well on large lesions and ensures full lesion capture but tends to over-segment, leading to poor performance on very small lesions. Individually, both Tversky Loss and Hausdorff Distance Loss performed worse than the baseline Dice Loss. However, when combined with other loss functions, they outperformed the baseline, suggesting that integrating both overlap-based and boundary-aware losses enhances segmentation performance.

This combined approach leverages the strengths of each loss function while mitigating their weaknesses, resulting in optimal performance. Tversky-HausdorffDT Loss achieves the highest Dice and NSD metrics, with the second-lowest MSD, whereas DiceFocal-HausdorffDT Loss has the lowest MSD, along with the second-highest Dice and NSD metrics. Overall, Tversky-HausdorffDT Loss performs best. These findings are supported by both evaluation metrics and qualitative analysis of segmentation masks on axial ADC maps.

Despite these improvements, the results only match the existing benchmark. However, a key observation is that our model reached this level of performance with significantly fewer training epochs. While compound loss functions require more epochs to converge compared to standalone losses, they still required far fewer epochs overall. In contrast, studies such as \cite{kazemi2025enhancing,koirala2025ensemble,aydin2025segresnet} trained for 400–1500 epochs using more advanced architectures to achieve only moderately better scores than ours.

\section{Conclusion}
In this study, we aimed to evaluate optimal loss function specifically for HIE lesion segmentation task and our proposed compound loss functions achieved the best Dice, MSD, and NSD metrics. However, despite achieving improvements with compound losses as hypothesized, this study has several limitations.

\subsection{Limitations}
\begin{itemize}
  \item None of the defined losses did well on extremely small lesions ( $<1 \%$ volume) due to high heterogeneity in the dataset and limited capability of the model itself.
  \item Only the validation set is used as a proxy for evaluation, as the test set is held out by the challenge organizers and has not been publicly released yet, following the conclusion of the challenge.
  \item Resampled label mask could introduce distortion in segmentation regions in comparison to the actual mask. A more robust approach would be to only resample maps to a fixed size and reverse resample the predicted binary masks to original size, spacing, direction, and origin for true comparison with the actual label mask.
\end{itemize}

\subsection{Future Work}
We’ve concluded that loss function optimization improves segmentation performance, specifically with compound losses. Future work can focus on evaluating the suggested compound losses with advanced architectures, data augmentation strategies, and ensemble and hybrid experimental setups.

While this study primarily investigates the optimization of loss functions to improve segmentation performance, the gains achieved are constrained by the limited dataset available. In scenarios with small datasets, training DL models from scratch can only go so far. Therefore, an alternative approach would be to avoid training altogether and instead use models that offer zero-shot generalizability. For example, MedSAM-2 \cite{zhu2024medical}, which is already fine-tuned on medical imaging, would be ideal. According to \cite{Bao2023}, we know that negative intensities in ZADC maps are correlated with lesion regions. Therefore, we can prompt MedSAM-2 with bounding boxes around negative intensities in ZADC maps and provide ADC maps as input for segmentation.

\bibliography{AI4CHL-arXiv}
\bibliographystyle{unsrtnat}

\end{document}